\documentclass[a4paper,fleqn]{cas-dc}

\usepackage[linesnumbered,ruled]{algorithm2e}

\usepackage{graphicx} 
\graphicspath{{figures/}} 
\usepackage{caption}
\usepackage{subcaption}
\usepackage{subcaption} 
\usepackage{amsmath,lineno,hyperref, graphicx, }
\usepackage{amssymb}
\modulolinenumbers[5]
\usepackage{float} 
\usepackage{lipsum}
\usepackage{verbatim}
\usepackage{multirow}
\usepackage{inputenc}
\usepackage[numbers]{natbib}
\usepackage{hyperref}
\usepackage[export]{adjustbox}
\usepackage{graphicx}
\usepackage{pdfpages}

\usepackage{color}

\begin{document}
\let\WriteBookmarks\relax
\def\floatpagepagefraction{1}
\def\textpagefraction{.001}
\shorttitle{Adaptive Large Neighborhood Search for CBPP}
\shortauthors{Kun He, et al.}
\newcommand{\etal}{{et al.}}
\title [mode = title]{Adaptive Large Neighborhood Search for Circle Bin Packing Problem}
\author[1]{Kun He}

\credit{Conceptualization of this study, Methodology, Software}

\address[1]{School of Computer Science and Technology, Huazhong University of Science and Technology, Wuhan 430074, China.}

\author[1]{Kevin Tole}
\fnmark[2]
\credit{Data curation, Writing - Original draft preparation}

\author[1]{Fei Ni}[%
]
\cormark[1]
\fnmark[3]

\author%
[1]
{Yong Yuan}
\cormark[1]
\fnmark[4]

\author%
[1]
{Linyun Liao}
\fnmark[5]

\cortext[cor1]{Corresponding author: Fei Ni, nifei@hust.edu.cn; Yong Yuan, m201873064@hust.edu.cn}



\begin{abstract}
        We address a new variant of packing problem called the circle bin packing problem (CBPP), which is to find a dense packing of circle items to multiple square bins so as to minimize the number of used bins. To this end, we propose an adaptive large neighborhood search (ALNS) algorithm, which uses our Greedy Algorithm with Corner Occupying Action (GACOA) to construct an initial layout. 
        The greedy solution is usually in a local optimum trap, and ALNS enables multiple neighborhood search that depends on the stochastic annealing schedule to avoid getting stuck in local minimum traps. 
        Specifically, ALNS perturbs the current layout to jump out of a local optimum by iteratively reassigns some circles and accepts the new layout with some probability during the search. The acceptance probability is adjusted adaptively using simulated annealing that fine-tunes the search direction in order to reach the global optimum. We benchmark computational results against GACOA in heterogeneous instances. ALNS always outperforms GACOA in improving the objective function, and in several cases, there is a significant reduction on the number of bins used in the packing. 
\end{abstract}

\begin{keywords}
	\texttt{NP-hard}\sep circle bin packing problem\sep adaptive large neighborhood search \sep simulated annealing \sep greedy heuristic 
\end{keywords}

\maketitle

\section{Introduction}
Packing problems form an important class of combinatorial optimization problems that have been well studied under numerous variants \cite{Specht,MOSTOFAAKBAR20061259,10.5555/1206604,Stracquadanio}. It is a classic type of NP-hard  problems, for which there is no deterministic algorithm to find exact solutions in polynomial time unless \emph{$P = NP$}. Also, there are numerous applications in the industry, such as shipping industry~\cite{THAPATSUWAN2012737,TOFFOLO2017526}, manufacturing materials \cite{WASCHER20071109,Maddaloni123}, advertisement placement \cite{Freund2004,DAWANDE2005}, loading problems \cite{BIRGIN200519,JUNQUEIRA201274}, and more exotic applications like origami folding \cite{Demaine2010CirclePF,An2018AnEA}.
Packing problems are well studied since 1832 Farkas \etal \cite{farkas,10.5555/1206604} investigated the occupying rate (density) of packing circle items in a bounded equilateral triangle bin, and since then tremendous improvements have been made~\cite{Graham95densepackings,lubachevsky2004dense}. 
In the past three decades, most researches focus on the single container packing. The container is either in square, circle, rectangle, or polygon shape \cite{LOPEZ2011512}, while the items can be rectangles, circles, triangles, or polygons. 

As one of the most classic packing problem, the circle packing problem (CPP) 
is mainly concerned with packing circular items in a container.
Researchers have proposed various methods for finding feasible near-optimal packing solutions \cite{hifi2007,cave2011,dosh,zeng2016iterated}, which fall into two types: constructive optimization approach and global optimization approach. 

The construction approach places the circle items one by one appropriately in the container based on a heuristic that defines the building rules to form a feasible solution. Most researches of this category either fix the position of the container's dimension and pack the items sequentially satisfying the constraints \cite{Dickinson2011}, or adjust the size of the container using a constructive approach \cite{zeng2016iterated}. Representative approaches include the Maximum Hole Degree (MHD) based algorithms \cite{Huang20031,HUANG2006,Huang2005}, among which Huang  \etal~\cite{Huang20031} came up with two greedy algorithms: "B.10" places the circle items based on MHD, while "B.15" strengthens the solution with a self-look-ahead search strategy. Another approach called Pruned Enriched Rosenbluth Method (PERM) \cite{LU20081742,PhysRevE.68.021113,PhysRevE.72.016704} is a population control algorithm incorporating the MHD strategy. There are also other heuristics such as the Best Local Position (BLP) based approaches~\cite{hifi2004,hifi2007,hifi2008,hifi2009}, which selects the best feasible positions to place the items among other positions that minimizes the size of the container.

On the other hand, the global optimization technique \cite{CAS} tries to solve the packing problem by improving the solution iteratively based on an initial solution, which is subdivided into two types. The first type is called the quasi-physical quasi-human algorithm \cite{quasi2016, quasi2017, HE201826}, which is mostly motivated by some physical phenomenon, or some wisdom observed in human activities~\cite{HE201567, He2016PackingUC}. The second type is called the meta-heuristic optimization, mainly built by defining an evaluation function that employs a trade-off of randomisation and local search that directs and re-models the basic heuristic to generate feasible solutions. The meta-heuristic searches an estimation in the solution space closing to the global optimum. Representative algorithms include the hybrid algorithm \cite{ZHANG20051941} that combines the simulated annealing and Tabu search \cite{glover1989tabu, glover1990tabu}. Recently another hybrid algorithm was proposed by combining Tabu search and Variable Neighborhood Descent, and yield state-of-the-art results~\cite{ZENG2018196}.


In this work, we address a new variant of packing problem called the two-dimensional circle bin packing problem (CBPP)~\cite{dosh}.
Given a collection of circles specified by their radii, we are asked to pack all items into a minimum number of identical square bins. A packing is called feasible if no circles overlap with each other or no circle be out of the bin boundary. The CBPP is a new type of geometric bin packing problem, and it is related to the well-studied 2D bin packing problem~\cite{LodiMV99,BansalLS-FOCS05}, which consists in packing a set of rectangular items into a minimum number of identical rectangular bins.  

This manuscript is an extended version with significant improvement on the algorithm of our previous conference publication~\cite{dosh}, among which we first introduce this problem and propose a Greedy Algorithm with Corner Occupying Action (GACOA) to construct a feasible dense layout~\cite{dosh}. In this paper, we further strengthen the packing quality and propose an Adaptive Large Neighborhood Search (ALNS) algorithm. ALNS first calls GACOA to construct an initial solution, then iteratively perturbs the current solution by randomly selecting any two used bins and unassigning circles that intersect a random picked region in each of the selected bin. Then we use GACOA to pack the outside circles back into the bin in order to form a complete solution.
The complete solution is accepted if the update layout increases the objective function or the decrease on the objective function is probabilistically allowed under the current annealing temperature. Note that the objective function is not the number of bins used but is defined to assist in weighing the performance to reach the global optimum of the new candidate solution. 
Computational numerical results show that ALNS always outperforms GACOA in improving the objective function, and sometimes ALNS even outputs packing patterns with less number of bins.

In this work, we make three main contributions: 

  1) we design a new form of objective function, embedding the number of containers used and the maximum difference between the containers with the highest density and the box with the lowest density. The new objective function can help identify the quality of the assignment, especially in the general case with the same number of bins.
  
  2) we propose a method for local search on the complete assignment solution. We select two bins randomly and generate a rectangular area for each bin with equal area. All the circles that intersect the rectangular area were unassigned and the remaining circles form the new partial solution. 
  
  3) we modify the conditions for receiving the new partial solution. The previous local neighborhood search algorithm only accepts new partial solutions with larger objective functions. However, it is not conductive to the global optimum to some extent. We apply the idea of simulated annealing to this new algorithm so that partial solutions with lower objective function values can also be accepted with a variable possibility.
  
The remaining of this paper is organised as follows. Section~\ref{sec:section2} introduces the mathematical constraints for the given problem, Section~\ref{sec:section3} presents the two frameworks used for the development of our algorithm. Section~\ref{sec:section4} further describes the objective function as well as the experimental setup. All the algorithms are computationally experimented and the results presented in Section~\ref{sec:section5}. Finally, Section~\ref{sec:section6} concludes with recommendations for future work.

\section{Problem Formulation}
\label{sec:section2}
Given a set of $n$ circles where item $C_{i}$ is in radius $r_i$ and $n$ identical square bins with side length $L$ (w.l.g. for any circle $C_{i}$, $2 \cdot r_i \leq L$), the CBPP problem is to locate the center coordinates of each $C_{i}$ such that any item is totally inside a container and there is no overlapping between any two items. The goal is to minimize the number of used bins, denoted as $K$ $(1\le K\le n)$. 

A feasible solution to the CBPP is a partition of the items into sets $\mathcal{S} = \langle S_{1}, S_{2}, \dots , S_{K} \rangle$ for the bins, and the packing constraints are satisfied in each bin. An optimal solution is the one in which $K$, the number of bins used, cannot be made any smaller. A summary of the necessary parameters is given in Table \ref{tab:f1}.

\begin{table}[width=1.0\linewidth,cols=2,pos=h]
\caption{Parameter regulation}
\centering
\begin{tabular*}{\tblwidth}{@{} |c|L|@{} }\hline
\toprule
Parameter & Description\\
\hline
$n$ & number of circles\tabularnewline
\hline 
$C_{i}$ &  the $i$-th circle\tabularnewline
\hline
$r_{i}$ &  radius of the $i$-th circle\tabularnewline
\hline 
$\left(x_{i},y_{i}\right)$ & center coordinates of $C_{i}$\tabularnewline
\hline 
$b_{k}$ & the bin that $C_{i}$ is assigned, $1\le k\le K$\tabularnewline
\hline 
$L$ & side length of square bin\tabularnewline
\hline 
$I_{ik}$ & indicator of the placement of $C_{i}$ into $b_{k}$\tabularnewline
\hline
$B_k$ & indicator of the use of bin $b_k$\tabularnewline
\hline
$d_{ij}$ & distance between $(x_{i},y_{i})$ and $(x_{j},y_{j})$\tabularnewline
\bottomrule

\end{tabular*}
\label{tab:f1}
\end{table}

Assume that the bottom left corner of each bin $b_{k}$ is placed at $(0,0)$ in it's own coordinate system. 
we formulate the CBPP as a constraint optimization problem.
\begin{equation}\textcolor{white}{.......................} 
\sum_{k=1}^{n} I_{ik}=1,\label{eq:1}
\end{equation} 
where 
\begin{equation}
I_{ik}\in \{0,1\}, ~~~i,k\in\{1,\ldots,n\}, 
\label{eq:2}
\end{equation} 
which implies that each circle is packed exactly once. Further, if a bin $b_{k}$ is used, then 
\begin{equation} 
B_{k}=\left\{ \begin{array}{ll} 1,\,\textrm{if}\,\sum_{i=1}^{n}I_{ik}>0,\,i,k\in \{1,\ldots, n\},\\0,\,\textrm{otherwise}.\end{array}\right.\label{eq:3} \end{equation} And, finally, for circles that are in the same bin, $I_{ik}=I_{jk}=1$, and $i,j,k\in \{1,\ldots,n\}$, no overlap is allowed, implying that
\begin{equation} 
d_{ij}=\sqrt{(x_{i}-x_{j})^{2}+(y_{i}-y_{j})^{2}}\ge (r_{i}+r_{j})I_{ik}I_{jk}.\label{eq:4} 
\end{equation}
%
Specifically, let the circles be ordered by their radii so that ${\it r_{1}}\ge  {\it r_{2}},  \ldots  \ge {\it r_{n}}$, $r_i \in \mathcal{R}^+$.  
To ensure that no item passes across the boundary of the bin, we ask that 
\begin{equation}
r_i \leq x_i \leq L-r_i, ~~ r_i \leq y_i \leq L-r_i
\label{eq:5}
\end{equation}
Conditions (\ref{eq:1})--(\ref{eq:4}) along with (\ref{eq:5}) are the constraints for CBPP. 

The overall goal of CBPP is to use as few bins as possible to pack the $n$ circles, which is 
\begin{equation} \textcolor{white}{.......................}\min K=\sum_{k=1}^{n}B_k.\label{eq:6} 
\end{equation}

\section{General Search Framework}
\label{sec:section3}

In this section, we introduce two general optimization search frameworks for constraint optimization problem that we will use for the development of our CBPP algorithm. The two frameworks are Large Neighborhood Search (LNS), and a variation of the well-studied simulated annealing process~\cite{Kirkpatrick}, Adaptive Large Neighborhood Search (ALNS)~\cite{gendreau2010handbook}. A particular advantage of ALNS is the capacity to move the iterated solution out of the local optimum.

\subsection{Large neighborhood search}

Large neighborhood search (LNS) is a technique to iteratively solve constraint optimization problems \cite{gendreau2010handbook,LARSSON2018103}. At each iteration, the goal is to find a more promising candidate solution to the problem, and traverse a better search path through the solution space. 
\newline\textbf{Definition 1. (Constraint optimization problem, COP)}. A constraint optimization problem $P=\left\langle n,D^{n},\mathcal{C},f\right\rangle $ is defined by an array of $n$ variables that can take values from a given domain $D^{n}$, subject to a set of constraints $\mathcal{C}$. $f$ is the objective function to measure the performance of assignment. An assignment is an array of values $\mathbf{a}^{n}\in D^{n}$. A constraint $c\in \mathcal{C}: D^{n} \rightarrow \{0,1\}$ is a predicate that decides whether an assignment $\mathbf{a}^{n}\in D^{n}$ is locally valid. A solution to $P$ is an assignment $\mathbf{a}^{n}$ that is locally valid for all constraints in $\mathcal{C}$, i.e. $c\left(\mathbf{a}^{n}\right)$ is true for all $c\in \mathcal{C}$. The optimal solution of $P$ is a solution that maximizes the objective function $f$.

In a nutshell, LNS starts from a non-optimal solution $\mathbf{a}^{n}$ and iteratively improves the solution until reaching an optimal or near-optimal solution.
The main ingredient in LNS is an effective algorithm for completing a partial solution.
\newline\textbf{Definition 2. (Partial solution)}. A partial solution $\left\langle k,\mathbf{a}^{k},I\right\rangle $ is an assignment $\mathbf{a}^{k}$ to a subset of $k$ variables (with their indexes $I$) of a constraint optimization problem $P$. Completing a partial solution means finding an assignment for the remaining variables $\left\{ a_{i}|i\notin I\right\} $ so that $\mathbf{a}^{n}$ is a solution to $P$.

At each iteration, LNS relaxes and repairs the solution by randomly
generating and then completing a partial solution (Alg.~\ref{alg:LNS-Algorithm}). If the new assignment $\mathbf{b}^n$ has higher objective function value, the previous assignment $\mathbf{a}^n$ will be replaced.

\begin{algorithm}
\SetKwInOut{Input}{Input}
\SetKwInOut{Output}{Output}
\SetAlgoLined
\SetKw{KwBreak}{break}
\Input{A COP $P=\left\langle n,D^{n},\mathcal{C},f\right\rangle $, \\
 number of iteration steps $N$}
\Output{A near-optimal solution $\mathbf{a}^{n}$}
$\mathbf{a}^{n}\leftarrow$
{compute\_initial\_solution}($P$) \\
\For{$i\leftarrow 1 $ \KwTo $N$}{
    $I\leftarrow $ generate\_partial\_solution($\mathbf{a}^{n}$)\;
    $\mathbf{b}^{n}\leftarrow $ 
    complete\_partial\_solution ($\mathbf{a}^{n},I,f$)\;
    \If{$f\left(\mathbf{b}^{n}\right)>f\left(\mathbf{a}^{n}\right)$}{
        $\mathbf{a}^{n}\leftarrow \mathbf{b}^{n}$\;
    }
}
\caption{Large neighborhood search}
\label{alg:LNS-Algorithm}
\end{algorithm}

\subsection{Adaptive large neighborhood search}

The new solution $\mathbf{b}^{n}$ obtained by complete\linebreak\_partial\_solution of LNS should be better than the previous solution $\mathbf{a}^{n}$ in order to be accepted (lines 5-7). But, in some sense, this approach actually limits the search for finding a global optimal solution. Thus, we consider an adaptive version of LNS, called ALNS~\cite{gendreau2010handbook}, obtained by altering LNS to allow for stepping to worse solutions. This depends on a predefined stochastic annealing schedule \cite{gendreau2010handbook,Kirkpatrick}, thus allowing for the solution search space to break out of local optima. We give a generic description of the ALNS algorithm in Alg.~\ref{alg:ALNS}, and a more complete explanation of the framework will be presented in  Section \ref{sec:section4}. 

\begin{algorithm}[h]
\SetKwInOut{Input}{Input}
\SetKwInOut{Output}{Output}
\SetAlgoLined
\SetKw{KwBreak}{break}
\Input{A COP $P=\left\langle n,D^{n},\mathcal{C},f\right\rangle$,\\
 number of iteration steps $N$}
\Output{A near-optimal solution $\mathbf{a}^{n}$ }
$\mathbf{a}^{n}\leftarrow $ compute\_initial\_solution($P$)\;
$\Theta \leftarrow$ initial temperature\;
$\theta \leftarrow \Theta$\; 
\For{$i\leftarrow 1 $ \KwTo $N$}{
    $I\leftarrow $ generate\_partial\_solution ($\mathbf{a}^{n}$)\;
    $\mathbf{b}^{n}\leftarrow $ 
    complete\_partial\_solution($\mathbf{a}^{n},I,f$)\;
    \If{\textrm{{acceptMove}}($\mathbf{b}^{n}, \mathbf{a}^{n}, \theta$)}{
        $\mathbf{a}^{n}\leftarrow \mathbf{b}^{n}$\;
    }
    $\theta = \theta - \Theta/N$\;
}
\caption{Adaptive large neighborhood search}
\label{alg:ALNS} 
\end{algorithm}

\vspace{-1em}
\section{ALNS for CBPP}
\label{sec:section4}

\subsection{Domain and objective function}
For the CBPP as a constraint optimization problem \linebreak (COP),  we define its domain $D^{n}$ as follows. For each circle $C_i$,  its assignment variables include $\langle x_{i},y_{i},I_{i1},...,I_{in} \rangle$. \linebreak We simplify the notation to $a_{ik} = \left\langle x_{i},y_{i},b_k\right\rangle$ if $I_{ik} = 1$. The corresponding domain $D= \mathbb{R}^{2}\times \left\{ 1,\ldots,n\right\}$ defines all possible assignments of a circle in the bin--coordinate space. Since each circle $C_{i}$ is constrained by an indicator function to be put only in a single bin, we abuse the notation slightly, simply use $a_{i}$ to denote $a_{ik}$, and place an emphasis on the circles rather than the containers, and each of the components where $i \in \{1,\ldots,n\}$ is a 3-tuple denoted as $\linebreak \left\langle x_{i},y_{i},b_k\right\rangle$. Thus, when referring to a solution to $P$, we write ${\mathbf a}^{n}$, and $D^{n}=D\times D\times \ldots \times D$ corresponds to the domain for the $n$ tuple variables. So ${\mathbf a}^{n}$ denotes a possible packing.

Let $L^2$ be the area of a bin, the density of $b_k$ of a packing is defined as 
\begin{equation}\textcolor{white}{.......................}
d_{k}({\mathbf a}^{n})=\frac{1}{L^2}\sum_{C_{i}\in S_{k}}\pi r_{i}^{2} I_{ik}.\label{eq:7}
\end{equation}
where $S_{k}$ is the set of items assigned to $b_k$.
Let $K$ be the number of used bins for a solution $\mathbf{a}^{n}$, so
\begin{equation} \textcolor{white}{.......................}
 K =\sum_{k=1}^{n}B_k.\label{eq:9} 
\end{equation}
%

%
\begin{equation*}
\begin{split}
\text{Let~~} d_{\min} &= \min \{d_{k}({\mathbf a}^{n})|1\le k\le K\},~~~~~~~~~~~~~~~~~~~~~~~~~~~~~~ \\
d_{\max} &= \max \{d_{k}({\mathbf a}^{n})|1\le k\le K\}. 
\end{split}
\end{equation*}
Here $d_{min}$ denotes the density of the sparsest bin and $d_{max}$ the density of the densest bin. 
We define a useful objective function, which will form part of our algorithm as 
\begin{equation}\textcolor{white}{..............}
f({\mathbf a}^{n})=-{K}+d_{max}-d_{min}.\label{eq:8}
\end{equation}

The larger the value of $f(\cdot)$ is, the better the packing is, since an increase in $f(\cdot)$ corresponds to a denser packing as circles move out of lower density bins.
In order to clarify the process for taking a complete candidate solution ${\mathbf a}^{n}$ for $P$ to a partial solution, the following formal definitions are required. 
%

%
%
%

%

Note that $0\le d_{max}-d_{min} \le1 $, this term is used for regularization. It implies that using fewer bins is preferable, that a difference in the number of bins is enough to compare two candidate solutions. With the same number of used bins in different solutions, we focus on the fullest bin and the emptiest bin on each candidate solution. The more dense the fullest bin is, the less wasted space is. The more sparse the emptiest bin is, the more concentrated the remaining still-reserved space is, which means it would be easier for assigning subsequent circles. So, the difference in density between the fullest bin and the emptiest bin determines the quality of each candidate solution.


\subsection{Construct initial solution}
\label{sec:cis} 

An initial solution can be quickly constructed by our greedy algorithm GACOA  (Alg.~\ref{alg:GACOA-Algorithm}).
\begin{equation} 
\text{compute\_initial\_solution}(P)= GACOA(L,\{{C_{i}}|1\leq i \leq n \})
\label{eq:12}
\end{equation}



For each circle, GACOA computes a set of candidate positions by greedily moving on to the next bin if a circle cannot be packed in any of the previous bins. In particular, each circle is packed according to the following criteria.
\newline\textbf{Definition 5. (Candidate packing position)}. 
A candidate-packing position of a circle in a bin is any position that places the circle tangent to a) any two packed items, or b) a packed item and the border of the bin, or c) two perpendicular sides of the bin (i.e. the corner).
\newline\textbf{Definition 6. (Feasible packing position)}. A packing position of a circle in a bin is feasible if it does not violate any constraints: circles do not
overlap and be fully contained in a bin. (See Eq. (\ref{eq:1})--(\ref{eq:5}) for detailed constraints). 
\newline\textbf{Definition 7. (Quality of packing position)}. The distance between the feasible packing position and the border of the bin is given by 
\begin{equation} q\left(x,y\right)= \left\{\min\left(d_{x},d_{y}\right),\max\left(d_{x},d_{y}\right)\right\}.\label{eq:7.1}
\end{equation}
where $d_{x}$ (resp. $d_{y}$) is a distance between the center of the circle and the closer side of the bin in the horizontal (resp. vertical) direction. 
For a circle in the current target bin, all feasible positions in the bin are sorted in dictionary order of $q(x,y)$.
The smaller, the better. 

We call an action that places a circle onto one of its candidate packing positions a Corner Occupying Action (COA). 
GACOA works by packing circles one by one in the decreasing order of their radii. Each circle considers the target bin in the increasing order of the bin index $k$. 
For bin $b_k$, a feasible candidate packing position with the highest quality is selected, i.e. a feasible assignment that maximizes $q(x,y)$ will be executed. Thus, the best feasible COA is selected that favours positions closer to the border of the bin. 
If there is no feasible assignment in bin $b_k$, we will try to pack in the next bin $b_{k+1}$. The pseudo code is given in Alg.~\ref{alg:GACOA-Algorithm}. 


\begin{algorithm}
\SetKwInOut{Input}{Input}
\SetAlgoLined
\SetKw{KwBreak}{break}
\Input{Bin side length $L$, 
a set of $n$ circles $\{C_i | 1\leq i\leq n\}$ with radii $r_1,\ldots,r_n$~ $(r_i \geq r_{i+1})$}
\KwResult{For each circle $C_i$, find a bin $b_k$, and place the circle center at $\left(x_i,y_i\right)$}
 $K\leftarrow 0$\
 
\For{$i\leftarrow 1$ \KwTo $n$}{
  \For{$k\leftarrow 1$ \KwTo $n$}{
    $S_{k}\leftarrow \emptyset$\; 
    
    \While{true}{
      \If{$k > K$}{$K = k$}  
      $S_{k}\leftarrow$ Feasible packing positions for $C_i$\;
      \If{$S_k \neq \emptyset$}{\KwBreak}
       $k \leftarrow k+1$\
    }
    A best packing position from $S_{k}$ is selected according $q(x,y)$\;
	$\left(x_i,y_i\right)\leftarrow \arg\max_{\left(x,y\right)\in S} q\left(x,y\right)$\;
	Execute this packing position to pack circle $C_i$ into bin $b_k$\;
	}
}
\caption{GACOA}\label{alg:GACOA-Algorithm}
\end{algorithm}

Partial solutions are generated from a complete solution by selecting
two bins at random and do perturbations. We randomly select a rectangular area of equal size in each bin, all circles that intersect the two rectangular areas will be taken out and added to the remain2assign set, and the complete solution becomes a partial solution.
The unassigned circles will be reassigned based on the partial solution. This is equivalent to perturbing a complete solution that has \linebreak reached a local optimum. 
Let function random\_ints$(m,M)$ returns $m$ distinct integers randomly selected from set $M$. Let function random\_real $(R)$ returns a random real number $0 < r\leq R$.

\begin{algorithm}
\SetKwInOut{Input}{Input}
\SetKwInOut{Output}{Output}
\SetAlgoLined
\SetKw{KwBreak}{break}
\Input{A (complete) solution $\mathbf{a}^n$}
\Output{A partial solution given by the indexes to keep $I$}
\tcp{Select two bins randomly}
$\left(k_1, k_2\right)\leftarrow$ random\_ints($2, \left\{1,\ldots,{K})\right\}$) \\ 
\tcp{Select a rectangular area in the first bin}
$rect_{1}\leftarrow\mathrm{sample\_rects}(b_{k_1})$  \\
\tcp{Select a rectangular area in the second bin}
$rect_{2}\leftarrow\mathrm{sample\_rects}(b_{k_2})$ \\
$remain2assign\leftarrow\bigcup_{j\in\left\{ 1,2\right\}}\{ i|\left\langle x_{i},y_{i},b_{k_j}\right\rangle \in\mathbf{a}^{n},I_{ik_j}=1 
~\bigwedge \mathbf {intersects}(C_i,rect_j)==True \} $\\
$I \leftarrow I / remain2assign$
\caption{Generate partial solution}
\label{alg:PartiaSolution-Algorithm}
\end{algorithm}

Alg.~\ref{alg:SampleRects-Algorithm} randomly selects a circle in the non-empty bin, and then randomly generates a rectangular area with the circle center be the center of the area. This guarantees that at least one circle will intersect the generated rectangular area (Here we simply use the envelope rectangle of the circle to check its intersection with the area). In most cases, more than one circle items intersect the rectangular area and will be unassigned at each iteration.
We choose to select two bins and generate one rectangular area for each bin. Only one bin can be sampled at a time, but when the partial solution and unassigned circle set are continued to be placed in the future, in the worst case, it will be put back as it is to obtain the same complete solution as before. The worst instance is that the previous partial solution did not leave enough free space to allocate the unassigned circles except for the generated area. If there is only one bin and one rectangular area, the unassigned circles are very likely to be put back into the previous generated rectangular area during the iteration, which means that this iteration process has no effect and does not help jump out of the local optimum. Thus, two bins are selected for each iteration so that the unassigned circles have more free space to be allocated. Even in the worst case, the algorithm will try to exchange the circles in the two rectangular areas, which ensures that there will be some disturbance per iteration. Of course, an alternative way is to sample only one bin and generate two rectangular areas for the bin, but two rectangular areas in different bins can increase the randomness. 





\begin{algorithm}
 \SetKwInOut{Input}{Input}
 \SetKwInOut{Output}{Output}
 \SetAlgoLined
 \SetKw{KwBreak}{break}
 \Input{Index of bin $k$; side length of bin $L$}
 \Output{rectangle area Rects}
 \tcp{Each rectangle is represented as a bottom-left point and a top-right point}
 $w\leftarrow\mathbf{random\_real}(L) $\tcp{the width of rectangle area is w}
 
 $h\leftarrow\mathbf{random\_real}(L) $ \tcp{the height of rectangle area is h}
 
 let $l_x \leftarrow 0$, $l_y \leftarrow 0$
 \tcp{ where $(l_x,l_y)$ is the coordinate of bottom-left point}
 
 \If{($ !b_k.empty()$) }{
  $i \leftarrow\mathbf{random\_ints}(1,\left\{ i\left|\left\langle x_{i},y_{i},b_{k}\right\rangle \in\mathbf{a}^{n}\right.\right\})$\\
  $l_x \leftarrow C_i.x-0.5w$\\
  $l_y \leftarrow C_i.y-0.5h$ 
 }
 $Rects =$ make\_ pair(point($l_x,l_y$),point($l_x+w,l_y+h$))
 \caption{\label{alg:SampleRects-Algorithm}sample\_rects}
\end{algorithm}

\begin{algorithm}
\SetKwInOut{Input}{Input}
\SetKwInOut{Output}{Output}
\SetAlgoLined
\SetKw{KwBreak}{break}
\Input{circle $C_i$, $rect_k$ with coordinate tuple\\
        <$l_x$, $l_y$, $l_x+w$, $l_y+h$>}
\Output{true or false}\tcp{returns if $C_i$ intersects the rectangle}
\If{$C_i.x-C_i.r\ge l_x+w$}{
return false \tcp{non-intersect}
}
\If{$C_i.y-C_i.r\ge l_y+h$}{
return false \tcp{non-intersect}
}
\If{$C_i.x+C_i.r\le l_x$}{
return false \tcp{non-intersect}
}
\If{$C_i.y+C_i.r\le l_y$}{
return false \tcp{non-intersect}
}
return true \tcp{intersect}
\caption{\label{alg:Intersects-Algorithm}intersects}
\end{algorithm}



\label{sec:gps}

\subsection{Complete a partial solution}
\label{sec:cps}

Completing a partial solution is performed efficiently by the {GACOA}
algorithm (Alg.~\ref{alg:GACOA-Algorithm}) restricted to the bins
$b_{k1},b_{k2}$ from which circles were unassigned in the previous
step. 
\begin{equation} 
\begin{aligned}
\textrm{{complete\_partial\_solution}}\left(\mathbf{a}^{n},I\right) =~~~~~~~~~~~~~~~~~~~~~~\\~~~~~\textrm{{GACOA}}\left(L,\left\{ C_{i}|i \in remain2assign\right\} \right)\label{eq:13} 
\end{aligned}
\end{equation}
%

$I$ is the partial solution generated by \linebreak generate\_partial\_solution(). GACOA is used to complete the partial solution.

\subsection{Acceptance metric}

\label{sec:acc}

A solution is accepted if it increases the objective function or
if the decrease in the objective function is probabilistically allowed
given the current annealing temperature $\theta$. This is the well-known simulated annealing move acceptance criteria \cite{Kirkpatrick},

\begin{equation}
\begin{aligned}
\textrm{{acceptMove}}(\mathbf{b}^n,\mathbf{a}^n,\theta) =
f(\mathbf{b}^n)>f(\mathbf{a}^n)~~~~~~~~~~~~~~\\
\vee{random\_real}(1)\leq e^{\frac{f(\mathbf{b}^{n})-f(\mathbf{a}^{n})}{\theta}}\label{eq:14}
\end{aligned}
\end{equation}


\subsection{ALNS for CBPP} 

The complete algorithm for solving CBPP requires various steps from the above. The ALNS procedure is started using the initial solution along with the temperature $\Theta$ and the number of iterations $N$. The initial solution is then broken using Alg.~\ref{alg:PartiaSolution-Algorithm} and re-completed using the new solution filled by GACOA. This procedure outputs a new candidate solution, which is then either accepted or rejected based on the acceptance metric with simulated annealing. At which point, if the iteration limit has not reached, the process restarts using the new solution as an input

\begin{figure*}[pos=H]
    \centering
    \includegraphics[width=\textwidth, scale=0.8]{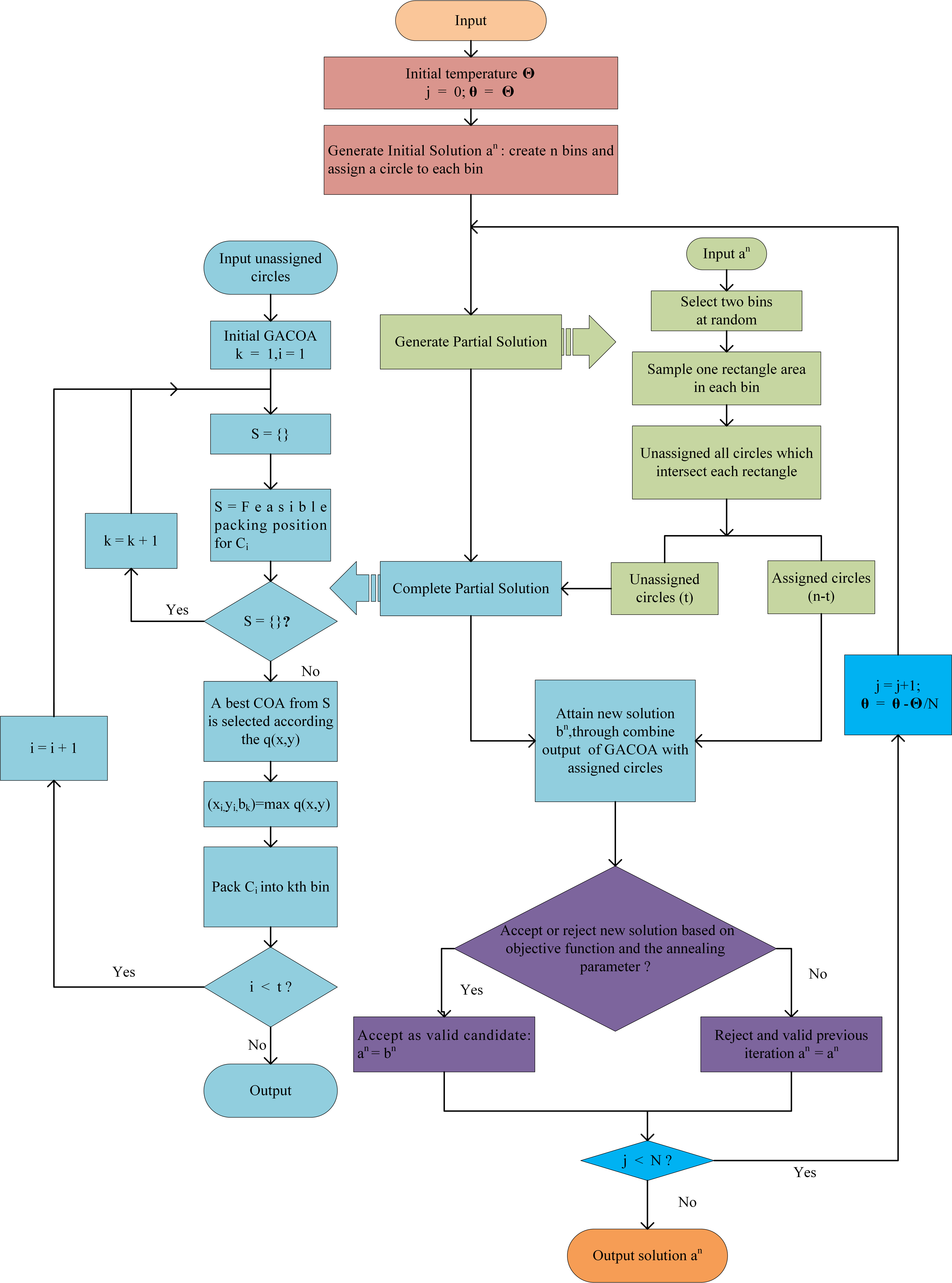}
    \caption{The parameter setup flow of ALNS.}  
    \label{fig:flow}
\end{figure*}

An illustration of the entire parameter setup flow of ALNS algorithm is shown in Figure~\ref{fig:flow}.

 \begin{figure*}[pos=H]
    \centering
    \includegraphics[scale=0.50]{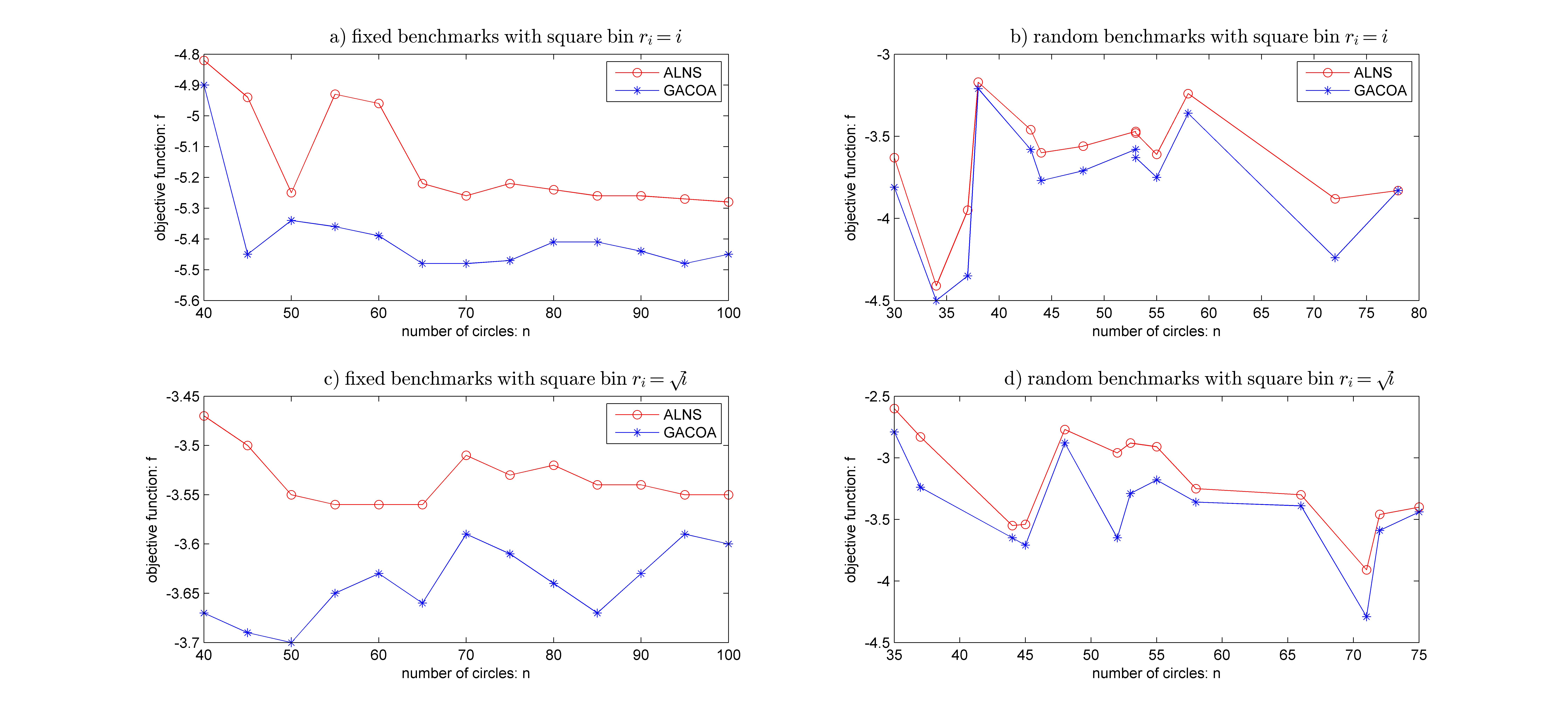}
    \caption{ALNS vs. GACOA.}
    \label{fig:my_label2}
\end{figure*}

\begin{table*}
\hfill{}
\begin{tabular}{|c|c||c||c|c|c|c|c|c||c|c|}
\hline 
$n_{0}$ & $n$ & alg. & bin 1 & bin 2 & bin 3 & bin 4 & bin 5 & bin 6 & $f$ & $f_{A}-f_{G}$\tabularnewline
\hline 
\hline 
\multirow{2}{*}{8} & \multirow{2}{*}{40} & {A} & 0.83 & 0.80 & 0.76 & 0.74 & 0.65 & - & -5.18 & \multirow{2}{*}{0.08}\tabularnewline
\cline{3-10} \cline{4-10} \cline{5-10} \cline{6-10} \cline{7-10} \cline{8-10} \cline{9-10} \cline{10-10} 
 &  & G & 0.83 & 0.74 & 0.71 & 0.76 & 0.73 & - & -5.10 & \tabularnewline
\hline 
\multirow{2}{*}{9} & \multirow{2}{*}{45} & {A} & 0.81 & 0.81 & 0.80 & 0.76 & 0.75 & - & -5.81 & \multirow{2}{*}{1.26}\tabularnewline
\cline{3-10} \cline{4-10} \cline{5-10} \cline{6-10} \cline{7-10} \cline{8-10} \cline{9-10} \cline{10-10} 
 &  & G & 0.76 & 0.72 & 0.75 & 0.75 & 0.72 & 0.21 & -6.55 & \tabularnewline
\hline 
\multirow{2}{*}{10} & \multirow{2}{*}{50} & A & 0.83 & 0.82 & 0.79 & 0.78 & 0.76 & 0.08 & -6.75 & \multirow{2}{*}{0.09}\tabularnewline
\cline{3-10} \cline{4-10} \cline{5-10} \cline{6-10} \cline{7-10} \cline{8-10} \cline{9-10} \cline{10-10} 
 &  & {G} & 0.84 & 0.78 & 0.80 & 0.75 & 0.72 & 0.18 & -6.66 & \tabularnewline
\hline 
\multirow{2}{*}{11} & \multirow{2}{*}{55} & A & 0.84 & 0.82 & 0.80 & 0.78 & 0.77 & - & -5.84 & \multirow{2}{*}{1.20}\tabularnewline
\cline{3-10} \cline{4-10} \cline{5-10} \cline{6-10} \cline{7-10} \cline{8-10} \cline{9-10} \cline{10-10} 
 &  & {G} & 0.83 & 0.78 & 0.77 & 0.75 & 0.69 & 0.19 & -6.64 & \tabularnewline
\hline 
\multirow{2}{*}{12} & \multirow{2}{*}{60} & A & 0.84 & 0.81 & 0.80 & 0.80 & 0.80 & - & -5.84 & \multirow{2}{*}{1.23}\tabularnewline
\cline{3-10} \cline{4-10} \cline{5-10} \cline{6-10} \cline{7-10} \cline{8-10} \cline{9-10} \cline{10-10} 
 &  & G & 0.84 & 0.79 & 0.78 & 0.72 & 0.69 & 0.23 & -6.61 & \tabularnewline
\hline 
\multirow{2}{*}{13} & \multirow{2}{*}{65} & A & 0.83 & 0.81 & 0.81 & 0.81 & 0.80 & 0.05 & -6.78 & \multirow{2}{*}{0.25}\tabularnewline
\cline{3-10} \cline{4-10} \cline{5-10} \cline{6-10} \cline{7-10} \cline{8-10} \cline{9-10} \cline{10-10} 
 &  & {G} & 0.83 & 0.79 & 0.76 & 0.72 & 0.69 & 0.31 & -6.52 & \tabularnewline
\hline 
\multirow{2}{*}{14} & \multirow{2}{*}{70} & A & 0.83 & 0.82 & 0.82 & 0.81 & 0.79 & 0.09 & -6.74 & \multirow{2}{*}{0.21}\tabularnewline
\cline{3-10} \cline{4-10} \cline{5-10} \cline{6-10} \cline{7-10} \cline{8-10} \cline{9-10} \cline{10-10} 
 &  & {G} & 0.84 & 0.82 & 0.77 & 0.71 & 0.71 & 0.32 & -6.52 & \tabularnewline
\hline 
\multirow{2}{*}{15} & \multirow{2}{*}{75} & A & 0.83 & 0.83 & 0.82 & 0.81 & 0.81 & 0.05 & -6.78 & \multirow{2}{*}{0.25}\tabularnewline
\cline{3-10} \cline{4-10} \cline{5-10} \cline{6-10} \cline{7-10} \cline{8-10} \cline{9-10} \cline{10-10} 
& & G & 0.85 & 0.81 & 0.76 & 0.70 & 0.70 & 0.32 & -6.53 & \tabularnewline
\hline 
\multirow{2}{*}{16} & \multirow{2}{*}{80} & A & 0.84 & 0.84 & 0.82 & 0.81 & 0.79 & 0.08 & -6.76 & \multirow{2}{*}{0.18}\tabularnewline
\cline{3-10} \cline{4-10} \cline{5-10} \cline{6-10} \cline{7-10} \cline{8-10} \cline{9-10} \cline{10-10} 
 &  & G & 0.85 & 0.82 & 0.81 & 0.73 & 0.71 & 0.26 & -6.58 & \tabularnewline
\hline 
\multirow{2}{*}{17} & \multirow{2}{*}{85} & A & 0.85 & 0.83 & 0.82 & 0.81 & 0.80 & 0.11 & -6.74 & \multirow{2}{*}{0.14}\tabularnewline
\cline{3-10} \cline{4-10} \cline{5-10} \cline{6-10} \cline{7-10} \cline{8-10} \cline{9-10} \cline{10-10} 
 &  & {G} & 0.85 & 0.82 & 0.79 & 0.75 & 0.75 & 0.26 & -6.60 & \tabularnewline
\hline 
\multirow{2}{*}{18} & \multirow{2}{*}{90} & A & 0.85 & 0.83 & 0.82 & 0.81 & 0.81 & 0.11 & -6.74 & \multirow{2}{*}{0.17}\tabularnewline
\cline{3-10} \cline{4-10} \cline{5-10} \cline{6-10} \cline{7-10} \cline{8-10} \cline{9-10} \cline{10-10} 
 &  & {G} & 0.85 & 0.80 & 0.80 & 0.79 & 0.71 & 0.29 & -6.57 & \tabularnewline
\hline 
\multirow{2}{*}{19} & \multirow{2}{*}{95} & A & 0.85 & 0.83 & 0.82 & 0.81 & 0.81 & 0.12 & -6.73 & \multirow{2}{*}{0.21}\tabularnewline
\cline{3-10} \cline{4-10} \cline{5-10} \cline{6-10} \cline{7-10} \cline{8-10} \cline{9-10} \cline{10-10} 
 &  & G & 0.84 & 0.81 & 0.80 & 0.77 & 0.69 & 0.32 & -6.52 & \tabularnewline
\hline 
\multirow{2}{*}{20} & \multirow{2}{*}{100} & A & 0.84 & 0.83 & 0.82 & 0.81 & 0.80 & 0.12 & -6.72 & \multirow{2}{*}{0.17}\tabularnewline
\cline{3-10} \cline{4-10} \cline{5-10} \cline{6-10} \cline{7-10} \cline{8-10} \cline{9-10} \cline{10-10} 
 &  & G & 0.86 & 0.81 & 0.78 & 0.77 & 0.71 & 0.31 & -6.55 & \tabularnewline
\hline 
\end{tabular}\hfill{}

\caption{Experimental results on the fixed benchmarks with square bins when $r_i = i$. The average improvement is 0.19. }
\label{tab:fixed-square}
\end{table*}

\begin{figure*}[pos=H]
 \centering
 \begin{subfigure}[b]{0.7\textwidth}
 \centering
  \includegraphics[width=\textwidth]{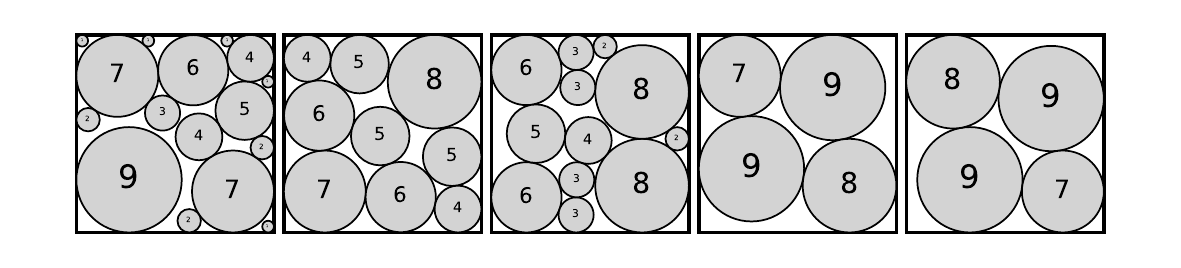}
  \caption{Packing layouts generated by ALNS algorithm}
  \label{fig:tae11}
 \end{subfigure}
 \begin{subfigure}[b]{0.7\textwidth}
 \centering
  \includegraphics[width=\textwidth]{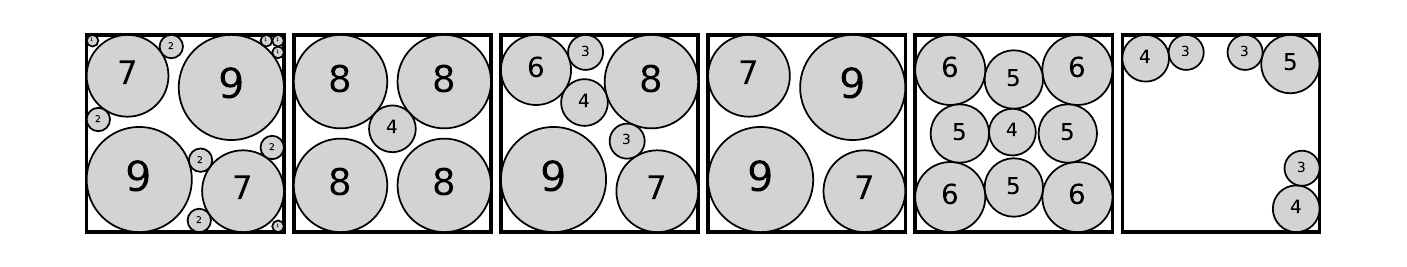}
  \caption{Packing layouts generated by GACOA algorithm}
  \label{fig:ptae41}
 \end{subfigure}
 \caption{Solution for the fixed benchmark when $n_{0}=9$ and $n=45$.}   
 \label{fig:layout}
\end{figure*}

\begin{table*}
\hfill{}%
\begin{tabular}{|c|c||c||c|c|c|c|c||c|c|}
\hline 
$n_{0}$ & \multicolumn{1}{c|}{$n$} & alg. & bin 1 & bin 2 & bin 3 & bin 4 & bin 5 & $f$ & $f_{A}-f_{G}$\tabularnewline
\hline 
\hline 
\multirow{2}{*}{8} & \multirow{2}{*}{34} & A & 0.83 & 0.80 & 0.74 & 0.73 & 0.24 & -4.41 & \multirow{2}{*}{0.09}\tabularnewline
\cline{3-9} \cline{4-9} \cline{5-9} \cline{6-9} \cline{7-9} \cline{8-9} \cline{9-9} 
 &  & G & 0.83 & 0.74 & 0.74 & 0.70 & 0.33 & -4.5 & \tabularnewline
\hline 
\multirow{2}{*}{9} & \multirow{2}{*}{30} & A & 0.81 & 0.80 & 0.77 & 0.44 & - & -3.63 & \multirow{2}{*}{0.18}\tabularnewline
\cline{3-9} \cline{4-9} \cline{5-9} \cline{6-9} \cline{7-9} \cline{8-9} \cline{9-9} 
 &  & G & 0.78 & 0.75 & 0.71 & 0.59 & - & -3.81 & \tabularnewline
\hline 
\multirow{2}{*}{10} & \multirow{2}{*}{37} & A & 0.81 & 0.80 & 0.79 & 0.76 & - & -3.95 & \multirow{2}{*}{0.4}\tabularnewline
\cline{3-9} \cline{4-9} \cline{5-9} \cline{6-9} \cline{7-9} \cline{8-9} \cline{9-9} 
 &  & G & 0.83 & 0.78 & 0.71 & 0.67 & 0.18 & -4.35 & \tabularnewline
\hline 
\multirow{2}{*}{11} & \multirow{2}{*}{38} & A & 0.83 & 0.81 & 0.79 & 0.00 & - & -3.17 & \multirow{2}{*}{0.04}\tabularnewline
\cline{3-9} \cline{4-9} \cline{5-9} \cline{6-9} \cline{7-9} \cline{8-9} \cline{9-9} 
 &  & G & 0.82 & 0.81 & 0.78 & 0.03 & - & -3.21 & \tabularnewline
\hline 
\multirow{2}{*}{12} & \multirow{2}{*}{43} & A & 0.83 & 0.81 & 0.80 & 0.29 & - & -3.46 & \multirow{2}{*}{0.12}\tabularnewline
\cline{3-9} \cline{4-9} \cline{5-9} \cline{6-9} \cline{7-9} \cline{8-9} \cline{9-9} 
 &  & G & 0.82 & 0.79 & 0.72 & 0.40 & - & -3.58 & \tabularnewline
\hline 
\multirow{2}{*}{13} & \multirow{2}{*}{44} & A & 0.83 & 0.81 & 0.78 & 0.43 & - & -3.6 & \multirow{2}{*}{0.17}\tabularnewline
\cline{3-9} \cline{4-9} \cline{5-9} \cline{6-9} \cline{7-9} \cline{8-9} \cline{9-9} 
 &  & G & 0.82 & 0.75 & 0.69 & 0.59 & - & -3.77 & \tabularnewline
\hline 
\multirow{2}{*}{14} & \multirow{2}{*}{48} & A & 0.84 & 0.82 & 0.80 & 0.40 & - & -3.56 & \multirow{2}{*}{0.15}\tabularnewline
\cline{3-9} \cline{4-9} \cline{5-9} \cline{6-9} \cline{7-9} \cline{8-9} \cline{9-9} 
 &  & G & 0.82 & 0.77 & 0.73 & 0.53 & - & -3.71 & \tabularnewline
\hline 
\multirow{2}{*}{15} & \multirow{2}{*}{53} & A & 0.85 & 0.83 & 0.81 & 0.32 & - & -3.47 & \multirow{2}{*}{0.11}\tabularnewline
\cline{3-9} \cline{4-9} \cline{5-9} \cline{6-9} \cline{7-9} \cline{8-9} \cline{9-9} 
 &  & G & 0.85 & 0.81 & 0.72 & 0.43 & - & -3.58 & \tabularnewline
\hline 
\multirow{2}{*}{16} & \multirow{2}{*}{55} & A & 0.85 & 0.82 & 0.81 & 0.46 & - & -3.61 & \multirow{2}{*}{0.14}\tabularnewline
\cline{3-9} \cline{4-9} \cline{5-9} \cline{6-9} \cline{7-9} \cline{8-9} \cline{9-9} 
 &  & G & 0.84 & 0.78 & 0.73 & 0.59 & - & -3.75 & \tabularnewline
\hline 
\multirow{2}{*}{17} & \multirow{2}{*}{53} & A & 0.84 & 0.82 & 0.80 & 0.32 & - & -3.48 & \multirow{2}{*}{0.15}\tabularnewline
\cline{3-9} \cline{4-9} \cline{5-9} \cline{6-9} \cline{7-9} \cline{8-9} \cline{9-9} 
 &  & G & 0.82 & 0.79 & 0.73 & 0.45 & - & -3.63 & \tabularnewline
\hline 
\multirow{2}{*}{18} & \multirow{2}{*}{72} & A & 0.84 & 0.83 & 0.83 & 0.72 & - & -3.88 & \multirow{2}{*}{0.36}\tabularnewline
\cline{3-9} \cline{4-9} \cline{5-9} \cline{6-9} \cline{7-9} \cline{8-9} \cline{9-9} 
 &  & G & 0.84 & 0.81 & 0.78 & 0.70 & 0.08 & -4.24 & \tabularnewline
\hline 
\multirow{2}{*}{19} & \multirow{2}{*}{58} & A & 0.83 & 0.83 & 0.81 & 0.07 & - & -3.24 & \multirow{2}{*}{0.12}\tabularnewline
\cline{3-9} \cline{4-9} \cline{5-9} \cline{6-9} \cline{7-9} \cline{8-9} \cline{9-9} 
 &  & G & 0.83 & 0.76 & 0.76 & 0.19 & - & -3.36 & \tabularnewline
\hline 
\multirow{2}{*}{20} & \multirow{2}{*}{78} & A & 0.84 & 0.84 & 0.81 & 0.67 & - & -3.83 & \multirow{2}{*}{0.01}\tabularnewline
\cline{3-9} \cline{4-9} \cline{5-9} \cline{6-9} \cline{7-9} \cline{8-9} \cline{9-9} 
 &  & G & 0.86 & 0.82 & 0.80 & 0.69 & - & -3.83 & \tabularnewline
\hline 
\end{tabular}\hfill{}

\caption{\label{tab:random-square}Experimental results on the random benchmarks for $r_i = i$. The average improvement is 0.12.}
\end{table*}

\begin{figure*}[pos=H]
 \centering
 \begin{subfigure}[b]{0.7\textwidth}
 \centering
  \includegraphics[width=\textwidth]{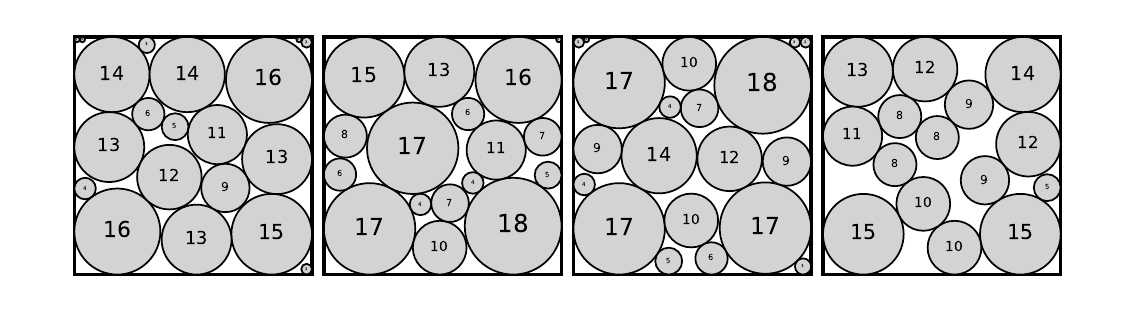}
  \caption{Packing layouts generated by ALNS algorithm}
  \label{fig:tae1}
 \end{subfigure}
 \begin{subfigure}[b]{0.7\textwidth}
 \centering
  \includegraphics[width=\textwidth]{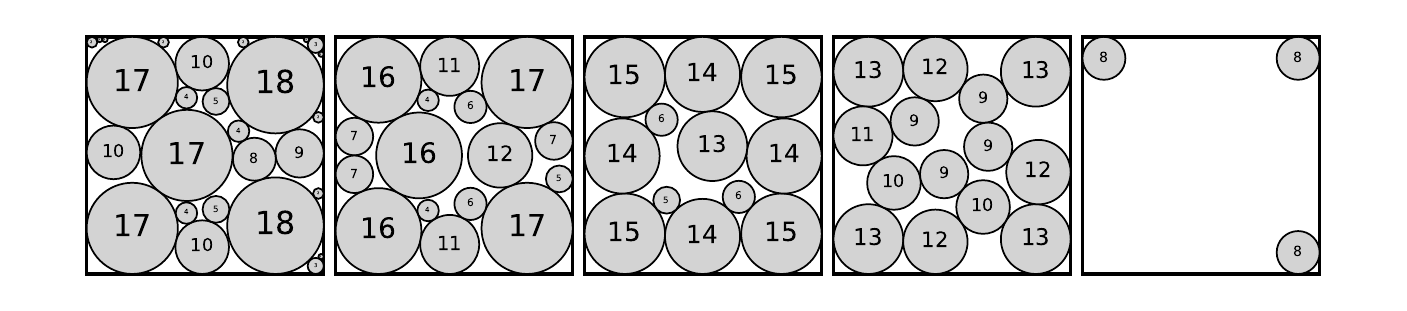}
  \caption{Packing layouts generated by GACOA algorithm}
  \label{fig:ptae42}
 \end{subfigure}
 \caption {\label{fig:layout18}} {Solution for the random benchmark  $n_{0}=18$ and $n=72$.} 
\end{figure*}

 %

\begin{table*}
\hfill{}%
\begin{tabular}{|c|c||c||c|c|c|c||c|c|}
\hline 
$n_{0}$ & $n$ & alg. & bin 1 & bin 2 & bin 3 & bin 4 & $f$ & $f_{A}-f_{G}$\tabularnewline
\hline 
\hline 
\multirow{2}{*}{8} & \multirow{2}{*}{40} & A & 0.83 & 0.83 & 0.76 & 0.30 & -3.47 & \multirow{2}{*}{0.21}\tabularnewline
\cline{3-8} \cline{4-8} \cline{5-8} \cline{6-8} \cline{7-8} \cline{8-8} 
 &  & G & 0.80 & 0.76 & 0.70 & 0.47 & -3.67 & \tabularnewline
 \hline
\multirow{2}{*}{9} & \multirow{2}{*}{45} & A & 0.83 & 0.82 & 0.79 & 0.33 & -3.5 & \multirow{2}{*}{0.19}\tabularnewline
\cline{3-8} \cline{4-8} \cline{5-8} \cline{6-8} \cline{7-8} \cline{8-8} 
 &  & G & 0.82 & 0.73 & 0.70 & 0.51 & -3.69 & \tabularnewline
 \hline 
 \multirow{2}{*}{10} & \multirow{2}{*}{50} & A & 0.83 & 0.81 & 0.80 & 0.38 & -3.55 & \multirow{2}{*}{0.15}\tabularnewline
\cline{3-8} \cline{4-8} \cline{5-8} \cline{6-8} \cline{7-8} \cline{8-8} 
 &  & G & 0.84 & 0.76 & 0.69 & 0.54 & -3.7 & \tabularnewline
 \hline
 \multirow{2}{*}{11} & \multirow{2}{*}{55} & A & 0.84 & 0.81 & 0.81 & 0.40 & -3.56 & \multirow{2}{*}{0.09}\tabularnewline
\cline{3-8} \cline{4-8} \cline{5-8} \cline{6-8} \cline{7-8} \cline{8-8} 
 &  & G & 0.83 & 0.78 & 0.77 & 0.48 & -3.65 & \tabularnewline
 \hline
 \multirow{2}{*}{12} & \multirow{2}{*}{60} & A & 0.83 & 0.83 & 0.81 & 0.39 & -3.56 & \multirow{2}{*}{0.07}\tabularnewline
\cline{3-8} \cline{4-8} \cline{5-8} \cline{6-8} \cline{7-8} \cline{8-8} 
 &  & G & 0.83 & 0.80 & 0.77 & 0.46 & -3.63 & \tabularnewline
 \hline
 \multirow{2}{*}{13} & \multirow{2}{*}{65} & A & 0.84 & 0.83 & 0.80 & 0.40 & -3.56 & \multirow{2}{*}{0.10}\tabularnewline
\cline{3-8} \cline{4-8} \cline{5-8} \cline{6-8} \cline{7-8} \cline{8-8} 
 &  & G & 0.83 & 0.82 & 0.72 & 0.49 & -3.66 & \tabularnewline
 \hline
 \multirow{2}{*}{14} & \multirow{2}{*}{70} & A & 0.85 & 0.82 & 0.81 & 0.36 & -3.51 & \multirow{2}{*}{0.08}\tabularnewline
\cline{3-8} \cline{4-8} \cline{5-8} \cline{6-8} \cline{7-8} \cline{8-8} 
 &  & G & 0.84 & 0.79 & 0.78 & 0.43 & -3.59 & \tabularnewline
 \hline
 \multirow{2}{*}{15} & \multirow{2}{*}{75} & A & 0.84 & 0.83 & 0.82 & 0.37 & -3.53 & \multirow{2}{*}{0.08}\tabularnewline
\cline{3-8} \cline{4-8} \cline{5-8} \cline{6-8} \cline{7-8} \cline{8-8} 
 &  & G & 0.85 & 0.80 & 0.75 & 0.46 & -3.61 & \tabularnewline
 \hline
 \multirow{2}{*}{16} & \multirow{2}{*}{80} & A & 0.86 & 0.83 & 0.81 & 0.38 & -3.52 & \multirow{2}{*}{0.12}\tabularnewline
\cline{3-8} \cline{4-8} \cline{5-8} \cline{6-8} \cline{7-8} \cline{8-8} 
 &  & G & 0.85 & 0.81 & 0.73 & 0.49 & -3.64 & \tabularnewline
 \hline
 \multirow{2}{*}{17} & \multirow{2}{*}{85} & A & 0.85 & 0.84 & 0.81 & 0.39 & -3.54 & \multirow{2}{*}{0.13}\tabularnewline
\cline{3-8} \cline{4-8} \cline{5-8} \cline{6-8} \cline{7-8} \cline{8-8} 
 &  & G & 0.83 & 0.79 & 0.77 & 0.50 & -3.67 & \tabularnewline
 \hline
 \multirow{2}{*}{18} & \multirow{2}{*}{90} & A & 0.87 & 0.82 & 0.80 & 0.41 & -3.54 & \multirow{2}{*}{0.09}\tabularnewline
\cline{3-8} \cline{4-8} \cline{5-8} \cline{6-8} \cline{7-8} \cline{8-8} 
 &  & G & 0.85 & 0.80 & 0.76 & 0.48 & -3.63 & \tabularnewline
 \hline
 \multirow{2}{*}{19} & \multirow{2}{*}{95} & A & 0.85 & 0.82 & 0.82 & 0.40 & -3.55 & \multirow{2}{*}{0.05}\tabularnewline
\cline{3-8} \cline{4-8} \cline{5-8} \cline{6-8} \cline{7-8} \cline{8-8} 
 &  & G & 0.86 & 0.82 & 0.77 & 0.45 & -3.59 & \tabularnewline
 \hline
 \multirow{2}{*}{20} & \multirow{2}{*}{100} & A & 0.86 & 0.82 & 0.82 & 0.41 & -3.55 & \multirow{2}{*}{0.05}\tabularnewline
\cline{3-8} \cline{4-8} \cline{5-8} \cline{6-8} \cline{7-8} \cline{8-8} 
 &  & G & 0.87 & 0.80 & 0.77 & 0.47 & -3.6 & \tabularnewline
 \hline
\end{tabular}\hfill{}

\caption{\label{tab:fixed-circle}Experimental results on the fixed benchmarks when $r_{i}=\sqrt{i}$. The average improvement of 0.22.}. 
\end{table*}
\begin{figure*}[pos=H]
	\centering
	\begin{subfigure}[b]{0.95\textwidth}
		\centering
		\includegraphics[width=\textwidth]{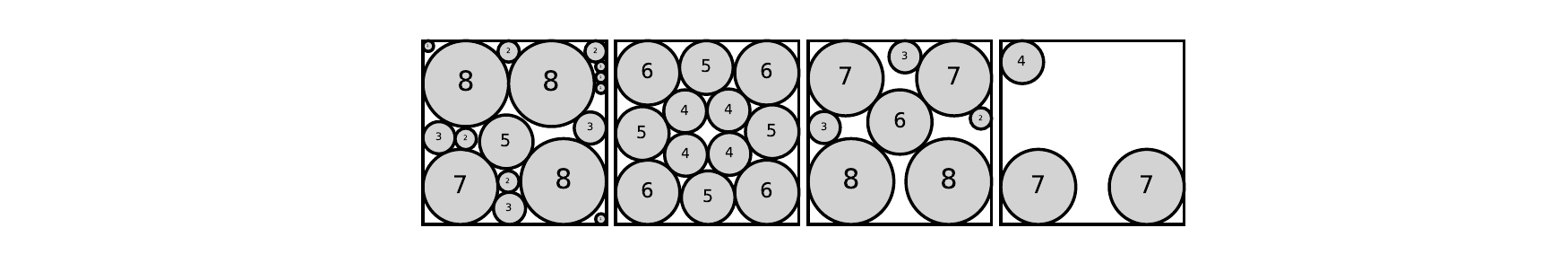}
		\caption{Packing layouts generated by ALNS algorithm}
		\label{fig:tae3}
	\end{subfigure}
	\begin{subfigure}[b]{0.95\textwidth}
		\centering
		\includegraphics[width=\textwidth]{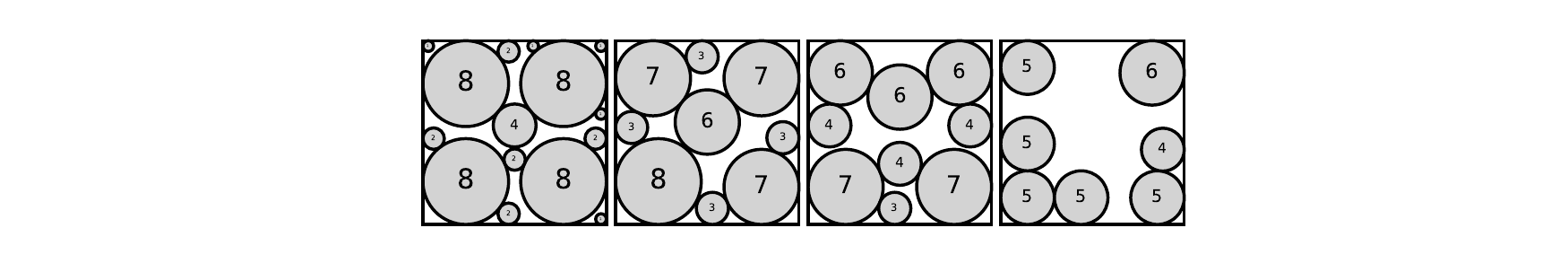}
		\caption{Packing layouts generated by GACOA algorithm}
		\label{fig:ptae}
	\end{subfigure}
	\caption{Solution for the fixed benchmark when $n_{0}=8$ and $n=40$ when $r_{i}=\sqrt{i}$.}   
	\label{fig:figxx}
\end{figure*}

 




\begin{table*}
\hfill{}%
\begin{tabular}{|c|c||c||c|c|c|c|c||c|c|}
\hline 
$n_{0}$ & $n$ & alg. & bin 1 & bin 2 & bin 3 & bin 4 & bin 5 & $f$ & $f_{A}-f_{G}$\tabularnewline
\hline 
\hline 
\multirow{2}{*}{8} & \multirow{2}{*}{37} & A & 0.84 & 0.83 & 0.67 & - & - & -2.83 & \multirow{2}{*}{0.41}\tabularnewline
\cline{3-9} \cline{4-9} \cline{5-9} \cline{6-9} \cline{7-9} \cline{8-9} \cline{9-9} 
 &  & G & 0.80 & 0.75 & 0.74 & 0.04 & - & -3.24 & \tabularnewline
\hline 
\multirow{2}{*}{9} & \multirow{2}{*}{35} & A & 0.82 & 0.81 & 0.42 & - & - & -2.6 & \multirow{2}{*}{0.19}\tabularnewline
\cline{3-9} \cline{4-9} \cline{5-9} \cline{6-9} \cline{7-9} \cline{8-9} \cline{9-9} 
 &  & G & 0.77 & 0.72 & 0.56 & - & - & -2.79 & \tabularnewline
\hline 
\multirow{2}{*}{10} & \multirow{2}{*}{45} & A & 0.83 & 0.82 & 0.81 & 0.37 & - & -3.54 & \multirow{2}{*}{0.17}\tabularnewline
\cline{3-9} \cline{4-9} \cline{5-9} \cline{6-9} \cline{7-9} \cline{8-9} \cline{9-9} 
 &  & G & 0.83 & 0.76 & 0.69 & 0.54 & - & -3.71 & \tabularnewline
\hline 
\multirow{2}{*}{11} & \multirow{2}{*}{44} & A & 0.84 & 0.81 & 0.81 & 0.39 & - & -3.55 & \multirow{2}{*}{0.10}\tabularnewline
\cline{3-9} \cline{4-9} \cline{5-9} \cline{6-9} \cline{7-9} \cline{8-9} \cline{9-9} 
 &  & G & 0.83 & 0.78 & 0.77 & 0.48 & - & -3.65 & \tabularnewline
\hline 
\multirow{2}{*}{12} & \multirow{2}{*}{52} & A & 0.84 & 0.82 & 0.80 & - & - & -2.96 & \multirow{2}{*}{0.69}\tabularnewline
\cline{3-9} \cline{4-9} \cline{5-9} \cline{6-9} \cline{7-9} \cline{8-9} \cline{9-9} 
 &  & G & 0.83 & 0.78 & 0.77 & 0.48 & - & -3.65 & \tabularnewline
\hline 
\multirow{2}{*}{13} & \multirow{2}{*}{71} & A & 0.85 & 0.83 & 0.81 & 0.76 & - & -3.91 & \multirow{2}{*}{0.38}\tabularnewline
\cline{3-9} \cline{4-9} \cline{5-9} \cline{6-9} \cline{7-9} \cline{8-9} \cline{9-9} 
 &  & G & 0.83 & 0.83 & 0.75 & 0.70 & 0.12 & -4.29 & \tabularnewline
\hline 
\multirow{2}{*}{14} & \multirow{2}{*}{55} & A & 0.84 & 0.83 & 0.75 & - & - & -2.91 & \multirow{2}{*}{0.27}\tabularnewline
\cline{3-9} \cline{4-9} \cline{5-9} \cline{6-9} \cline{7-9} \cline{8-9} \cline{9-9} 
 &  & G & 0.85 & 0.80 & 0.76 & 0.03 & - & -3.18 & \tabularnewline
\hline 
\multirow{2}{*}{15} & \multirow{2}{*}{53} & A & 0.83 & 0.83 & 0.71 & - & - & -2.88 & \multirow{2}{*}{0.41}\tabularnewline
\cline{3-9} \cline{4-9} \cline{5-9} \cline{6-9} \cline{7-9} \cline{8-9} \cline{9-9} 
 &  & G & 0.84 & 0.74 & 0.65 & 0.13 & - & -3.29 & \tabularnewline
\hline 
\multirow{2}{*}{16} & \multirow{2}{*}{48} & A & 0.83 & 0.82 & 0.60 & - & - & -2.77 & \multirow{2}{*}{0.11}\tabularnewline
\cline{3-9} \cline{4-9} \cline{5-9} \cline{6-9} \cline{7-9} \cline{8-9} \cline{9-9} 
 &  & G & 0.81 & 0.75 & 0.69 & - & - & -2.88 & \tabularnewline
\hline 
\multirow{2}{*}{17} & \multirow{2}{*}{72} & A & 0.84 & 0.82 & 0.81 & 0.30 & - & -3.46 & \multirow{2}{*}{0.13}\tabularnewline
\cline{3-9} \cline{4-9} \cline{5-9} \cline{6-9} \cline{7-9} \cline{8-9} \cline{9-9} 
 &  & G & 0.82 & 0.77 & 0.77 & 0.41 & - & -3.59 & \tabularnewline
\hline 
\multirow{2}{*}{18} & \multirow{2}{*}{66} & A & 0.82 & 0.81 & 0.80 & 0.12 & - & -3.3 & \multirow{2}{*}{0.09}\tabularnewline
\cline{3-9} \cline{4-9} \cline{5-9} \cline{6-9} \cline{7-9} \cline{8-9} \cline{9-9} 
 &  & G & 0.82 & 0.78 & 0.75 & 0.21 & - & -3.39 & \tabularnewline
\hline 
\multirow{2}{*}{19} & \multirow{2}{*}{75} & A & 0.84 & 0.82 & 0.81 & 0.24 & - & -3.4 & \multirow{2}{*}{0.04}\tabularnewline
\cline{3-9} \cline{4-9} \cline{5-9} \cline{6-9} \cline{7-9} \cline{8-9} \cline{9-9} 
 &  & G & 0.84 & 0.81 & 0.78 & 0.28 & - & -3.44 & \tabularnewline
\hline 
\multirow{2}{*}{20} & \multirow{2}{*}{58} & A & 0.81 & 0.81 & 0.79 & 0.06 & - & -3.25 & \multirow{2}{*}{0.10}\tabularnewline
\cline{3-9} \cline{4-9} \cline{5-9} \cline{6-9} \cline{7-9} \cline{8-9} \cline{9-9} 
 &  & G & 0.78 & 0.78 & 0.76 & 0.14 & - & -3.36 & \tabularnewline
\hline 
\hline 
\end{tabular}\hfill{}

\caption{\label{tab:random-circle2}Experimental results on the random benchmarks for $r_{i}=\sqrt{i} $. The average improvement is 0.23.}
\end{table*}

\begin{figure*}[pos=H]
 \centering
 \begin{subfigure}[b]{0.7\textwidth}
 \centering
  \includegraphics[width=\textwidth]{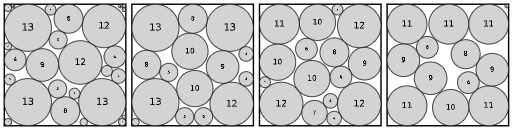}
  \caption{Packing layouts generated by ALNS algorithm}
  \label{fig:tae4}
 \end{subfigure}
 \begin{subfigure}[b]{0.7\textwidth}
 \centering
  \includegraphics[width=\textwidth]{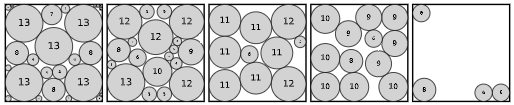}
  \caption{Packing layouts generated by GACOA algorithm}
  \label{fig:ptae4}
 \end{subfigure}
\caption {\label{fig:figxx2}} {Solution for the random benchmark with $n_{0}=13$ and $n=71$ when $r_{i}=\sqrt{i}$.}  
\end{figure*}

 %
 ~ 

\begin{table*}[pos=H]
\hfill{}%
\begin{tabular}{|c|c|c|c|c|c|c|c|c|}
\hline 
 & \multicolumn{4}{c|}{$r_i = i$} & \multicolumn{4}{c|}{$r_{i}=\sqrt{i} $}\tabularnewline
\hline 
\hline 
 & \multicolumn{2}{c|}{fixed} & \multicolumn{2}{c|}{random} & \multicolumn{2}{c|}{fixed} & \multicolumn{2}{c|}{random}\tabularnewline
\hline 
 $n_0$ & $n$ & $t$ & $n$ & $t$ & $n$ & $t$ & $n$ & $t$\tabularnewline
\hline 
8 & 40 & 82 & 34 & 59 & 40 & 53 & 37 & 42\tabularnewline
\hline 
9 & 45 & 72 & 30 & 63 & 45 & 87 & 35 & 61\tabularnewline
\hline 
10 & 50 & 74 & 37 & 69 & 50 & 76 & 45 & 59\tabularnewline
\hline 
11 & 55 & 91 & 38 & 160 & 55 & 88 & 44 & 82\tabularnewline
\hline 
12 & 60 & 103 & 43 & 120 & 60 & 107 & 52 & 81\tabularnewline
\hline 
13 & 65 & 116 & 44 & 127 & 65 & 137 & 71 & 142\tabularnewline
\hline 
14 & 70 & 135 & 48 & 148 & 70 & 152 & 55 & 145\tabularnewline
\hline 
15 & 75 & 154 & 53 & 177 & 75 & 182 & 53 & 210\tabularnewline
\hline 
16 & 80 & 179 & 55 & 189 & 80 & 180 & 48 & 163\tabularnewline
\hline 
17 & 85 & 204 & 53 & 179 & 85 & 202 & 72 & 227\tabularnewline
\hline 
18 & 90 & 222 & 72 & 339 & 90 & 216 & 66 & 197\tabularnewline
\hline 
19 & 95 & 247 & 58 & 209 & 95 & 234 & 75 & 261\tabularnewline
\hline 
20 & 100 & 279 & 78 & 366 & 100 & 255 & 58 & 251\tabularnewline
\hline 
\end{tabular}\hfill{}
\caption{Runtimes for ALNS execution in all benchmarks. }
\label{tab:Runtime}
\end{table*}

\section{Computational Experiments}
\label{sec:section5}

To evaluate the competency of the proposed approach, we implemented the ALNS for CBPP using the Visual C++ programming language. All results were generated by setting $N=2\times10^{6}$ (Alg.~\ref{alg:ALNS}) and obtained in a computer with Intel Core i7-8550U CPU @ 1.80GHz. 



We generated benchmarks based on two groups of instances downloaded from  \url{www.packomonia.com} for Single Circle Packing Problem (SCCP): $r_i=i$ for wide variation instances and $r_{i}=\sqrt{i}$ for smaller variation instances. On the packomania website, we use the range seed (number of circles for SCCP) from 8 to 20, and  
generate sets of benchmarks as follows: 
For each square bin, the best solution found in \cite{packomania} for the SCPP was used to fix the bin size $L$ as the best known record ${L_{best~known~record}}$ and we generate two sets of benchmarks called ``fixed" and ``random".Let $S=\{ C_{i}|1\leq i$ $\leq n \}$ be the set of circles packed in the current best solution for SCPP . The fixed set of CBPP benchmarks contains exactly $5$ copies of each circle packing for SCPP. The rand set of CBPP benchmarks contains a random number ($2\leq r\leq5$) of copies of each circle (i.e., the number of copies of each circle varies across the same benchmark) packed for SCPP.

In the following tables we get $52 \times 2$ generated instances from the two groups of instances, we compare the number of bins and the objective value for the best solution obtained by our search algorithm ALNS with the solution obtained from our constructive algorithm GACOA. They contain, for each original number of circles ($n_{0}$) and actual number of replicated circles in the CBPP benchmark ($n$), two rows showing the bin densities for ALNS (A) and GACOA (G) algorithm. The last two columns contain the final value of the objective function obtained in each algorithm and the relative improvement of ALNS over GACOA. Figure~\ref{fig:my_label2} shows in depth the typical behavior in comparison of ALNS and GACOA on the two benchmark instances. The objective function ($f$) in the y-axis under the number of circles in the x-axis. We can observe the packing occupying rate of ALNS is higher than that of GACOA. An in-depth analysis is explained in Section~\ref{sec:section5.1} and Section~\ref{sec:section5.2}.


\subsection{Comparison on $r=i$}
\label{sec:section5.1}
We run ALNS and GACOA on the benchmark instances of $r=i$. The number of unequal circle items ranges from 8 to 20 seeds for both fixed and random square settings. Table \ref{tab:fixed-square} shows the computational results of the fixed copies of 5 for each $n_0$. Table \ref{tab:random-square} displays the random number of copies of the circle items ranging from  $2\leq r\leq5$ of the corresponding seed. From Table \ref{tab:fixed-square} we can observe that when $n=45, 55,$ or $ 60$ on the fixed benchmark, ALNS utilises 5 bins, while GACOA uses 6 bins to pack the circle items. Figure \ref{fig:layout} illustrates the packing layout for $n=45$. On the other hand, for the random instances in Table \ref{tab:random-square} when $n=37$ or $72$, ALNS also minimizes the number of bins by packing circles in 4 bins while GACOA packs in 5 bins. Figure \ref{fig:layout18}  illustrates the packing layout for $n=72$. In summary, from the results of all the instances, ALNS returns feasible results, and has an overall average improvement of $19\%$ for the fixed benchmarks and $12\%$ for the random benchmarks when compared to GACOA.

\subsection{Comparison on $r_{i}=\sqrt{i}$}
\label{sec:section5.2}
 Similarly we also run ALNS and GACOA on the benchmark instances of $r_{i}=\sqrt{i}$, which contains a smaller variation of radii. Table \ref{tab:fixed-circle} contains fixed instances while Table \ref{tab:random-circle2} contains random instances. The instances also range from 8 to 20 seeds. As an illustration, we select $n_0=8$ and $n=40$ from Table \ref{tab:fixed-circle} and illustrate the layout in Figure \ref{fig:figxx}. The occupying rate for the first three bins is higher for ALNS when compared to that of GACOA, showing that most circle items have maximally occupied each bin's area. The fourth last bin's density of the packed items for ALNS is lower than that of GACOA, indicating that ALNS contains fewer packed circle items in the last bin. On the random benchmarks in Table \ref{tab:random-circle2} for $n_0$ at instance 8, 12, 13, 14 and 15, we also observe that ALNS uses one lesser bin in total when compared to GACOA. The average improvement of ALNS over GACOA is $22\%$ for the fixed instances and $23\%$ for the random instances, respectively. 

 \subsection{Further discussion}
  The run-times for ALNS at each benchmark are shown in Table \ref{tab:Runtime} (column of $t$ in seconds). We see that ALNS computes the instances efficiently in less than 300 seconds for 100 items. By comparison, as a greedy algorithm, GACOA completes the calculation in microseconds on any of these benchmarks. Such property facilitates the high efficiency of the ALNS algorithm. 
  
  In summary, for all the generated instances, we compare the number of bins and the objective value for the best solution obtained by ALNS algorithm with solutions obtained by GACOA. The results clearly show that ALNS consistently improves the objective function value as compared to GACOA over all sets of benchmarks, and it was even able to reduce the number of bins used in some benchmarks.\newpage The results show that our objective function does guide the ALNS algorithm to search for dense packing and promote reducing the number of bins used.

\section{Conclusions}
\label{sec:section6}
  We address a new type of packing problem, two dimensional circle bin packing problem (2D-CBPP), and propose an adaptive local search algorithm for solving this NP-Hard problem. The algorithm adopts a simulated annealing search on our greedy constructive algorithm.
  The initial solution is built by the greedy algorithm. Then during the search, we generate a partial solution by randomly selecting  rectangular areas in two bins and remove the circle items that intersect the areas. And we implement our greedy algorithm for completing partial solutions during the search. To facilitate the search, we design a new form of objective function, embedding the number of containers used and the maximum gap of the densities of different containers. A new solution is conditionally accepted by simulated annealing, completing one iteration of the search.
  
   Despite to all the improvements in this work, it is highly noted that the proposed problem is indeed challenging for combinatorial optimization heuristics and future researches are needed to get better solutions and generate high quality benchmarks. Implementing an adaptive local neighborhood search seems to be an attractive meta-heuristic to adopt. we would like to explore the idea to other circle bin packing problems, and extend our approach in addressing three dimensional circle bin packing problem which is more challenging with many applications that deserves proper attention.

\section*{Acknowledgement}
    This work was supported by the National Natural Science Foundation of China (Grant No. U1836204, U1936108) and the Fundamental Research Funds for the Central Universities  (2019kfyXKJC021).
    
\bibliographystyle{elsarticle-num}
\bibliography{mybibfile.bib}

\begin{thebibliography}{10}
\expandafter\ifx\csname url\endcsname\relax
  \def\url#1{\texttt{#1}}\fi
\expandafter\ifx\csname urlprefix\endcsname\relax\def\urlprefix{URL }\fi
\expandafter\ifx\csname href\endcsname\relax
  \def\href#1#2{#2} \def\path#1{#1}\fi

\bibitem{Specht}
E.~Specht, A precise algorithm to detect voids in polydisperse circle packings,
  Proceedings of the Royal Society of London Series A 471 (2015) 19.

\bibitem{MOSTOFAAKBAR20061259}
M.~M. Akbar, M.~S. Rahman, M.~Kaykobad, E.~Manning, G.~Shoja, Solving the
  multidimensional multiple-choice knapsack problem by constructing convex
  hulls, Computers \& Operations Research 33 (2006).

\bibitem{10.5555/1206604}
S.~P. G., M.~C. Markot, T.~Csendes, E.~Specht, L.~G. Casado, I.~Garcia~a, New
  Approaches to Circle Packing in a Square: With Program Codes (Springer
  Optimization and Its Applications), Springer-Verlag, 2007.

\bibitem{Stracquadanio}
S.~G, Greco.O, Conca.P, Cutello.V, P.~M, N.~G, Packing equal disks in a unit
  square: an immunological optimization approach, in: International Workshop on
  Artificial Immune Systems (AIS), 2015, pp. 1--5.

\bibitem{THAPATSUWAN2012737}
P.~Thapatsuwan, P.~Pongcharoen, C.~Hicks, W.~Chainate, Development of a
  stochastic optimisation tool for solving the multiple container packing
  problems, International Journal of Production Economics 140 (2012) 737 --
  748.

\bibitem{TOFFOLO2017526}
T.~A. Toffolo, E.~Esprit, T.~Wauters, G.~V. Berghe, A two-dimensional heuristic
  decomposition approach to a three-dimensional multiple container loading
  problem, European Journal of Operational Research 257 (2017) 526 -- 538.

\bibitem{WASCHER20071109}
G.~Wäscher, H.~Haußner, H.~Schumann, An improved typology of cutting and
  packing problems, European Journal of Operational Research 183 (2007) 1109 --
  1130.

\bibitem{Maddaloni123}
M.~A, Colla.V, Nastasi.G, S.~M., I.~V, A bin packing algorithm for steel
  production, 2016, pp. 19--24.

\bibitem{Freund2004}
A.~Freund, J.~S. Naor, Approximating the advertisement placement problem,
  Journal of Scheduling 7 (2004) 365--374.

\bibitem{DAWANDE2005}
M.~Dawande, S.~Kumar, C.~Sriskandarajah, A note on the minspace problem,
  Journal of Scheduling 8 (2005) 97--106.

\bibitem{BIRGIN200519}
B.~E. G, M.~J.M.z, R.~D.Pn, Optimizing the packing of cylinders into a
  rectangular container: A nonlinear approach, European Journal of Operational
  Research 160 (2005) 19 -- 33.

\bibitem{JUNQUEIRA201274}
L.~Junqueira, R.~Morabito, D.~S. Yamashita, Three-dimensional container loading
  models with cargo stability and load bearing constraints, Computers \&
  Operations Research 39 (2012) 74 -- 85.

\bibitem{Demaine2010CirclePF}
M.~L. Demaine, S.~P. Fekete, R.~J. Lang, Circle packing for origami design is
  hard, ArXiv abs/1008.1224 (2010).

\bibitem{An2018AnEA}
B.~An, S.~Miyashita, A.~Ong, M.~T. Tolley, M.~L. Demaine, E.~D. Demaine, R.~J.
  Wood, D.~Rus, An end-to-end approach to self-folding origami structures, IEEE
  Transactions on Robotics 34 (2018) 1409--1424.

\bibitem{farkas}
F.~Bolyai, Wolfgangi bolyai de bolya: Tentamen iuventutem studiosam in elementa
  math-eseos purae elementaris ac sublimioris methodo intuitiva evidentiaque
  huic propria introducendi,cum appendice triplici. in latin.
  marosvasarhelyini,second edition in 1904 2 (1832-33) 119--122.

\bibitem{Graham95densepackings}
R.~L. Graham, B.~D. Lubachevsky, Dense packings of equal disks in an
  equilateral triangle: From 22 to 34 and beyond, J. of Combinatorics 2 (1995)
  pp. (electronic).

\bibitem{lubachevsky2004dense}
R.~L. G. B.~D. Lubachevsky, Dense packings of equal disks in an equilateral
  triangle: From 22 to 34 and beyond, The Electronic Journal of Combinatorics 2
  (1995), A1 (2004) 39.

\bibitem{LOPEZ2011512}
C.~López, J.~Beasley, A heuristic for the circle packing problem with a
  variety of containers, European Journal of Operational Research 214 (2011)
  512 -- 525.

\bibitem{hifi2007}
M.~Hifi, R.~M’Hallah, A dynamic adaptive local search algorithm for the
  circular packing problem, European Journal of Operational Research 183 (2007)
  1280 -- 1294.

\bibitem{cave2011}
K.~He, W.~Huang, An efficient placement heuristic for three-dimensional
  rectangular packing, Computers \& Operations Research 38 (2011) 227 -- 233.

\bibitem{dosh}
K.~He, M.~Dosh, A greedy heuristic based on corner occupying action for the 2d
  circular bin packing problem, Springer Singapore (2017) 75 -- 85.

\bibitem{zeng2016iterated}
Z.~Zeng, X.~Yu, K.~He, W.~Huang, Z.~Fu, Iterated tabu search and variable
  neighborhood descent for packing unequal circles into a circular container,
  European Journal of Operational Research 250 (2016) 615--627.

\bibitem{Dickinson2011}
W.~Dickinson, D.~Guillot, A.~Keaton, S.~Xhumari, Optimal packings of up to five
  equal circles on a square flat torus, Contributions to Algebra and Geometry
  52 (2011) 315--333.

\bibitem{Huang20031}
W.~Huang, Y.~Li, B.~Jurkowiak, C.~Li, R.~Xu, A two-level search strategy for
  packing unequal circles into a circle container, in: Principles and Practice
  of Constraint Programming, Springer Berlin Heidelberg, 2003, pp. 868--872.

\bibitem{HUANG2006}
W.~Huang, Y.~Li, C.~Li, R.~Xu, New heuristics for packing unequal circles into
  a circular container, Computers \& Operations Research 33 (2006) 2125 --
  2142.

\bibitem{Huang2005}
W.~Huang, Y.~Li, H.~Akeb, C.~Li, Greedy algorithms for packing unequal circles
  into a rectangular container, Journal of the Operational Research Society 56
  (2005) 539--548.

\bibitem{LU20081742}
Z.~Lü, W.~Huang, Perm for solving circle packing problem, Computers \&
  Operations Research 35 (2008) 1742 -- 1755.

\bibitem{PhysRevE.68.021113}
H.~P. Hsu, V.~Mehra, W.~Nadler, G.~Peter, Growth based optimization algorithm
  for lattice heteropolymers, Phys. Rev. E 68 (2003) 021113.

\bibitem{PhysRevE.72.016704}
W.~Huang, Z.~P. Lu, H.~Shi, Growth algorithm for finding low energy
  configurations of simple lattice proteins, Physics Review E 72 (2005) 016704.

\bibitem{hifi2004}
M.~Hifi, R.~MHallah, Approximate algorithms for constrained circular cutting
  problems, Computers \& Operations Research 31 (2004) 675 -- 694.

\bibitem{hifi2008}
M.~Hifi, R.~M’Hallah, Adaptive and restarting techniques based algorithms for
  circular packing problems, Computational Optimization and Applications 39
  (2008) 17--35.

\bibitem{hifi2009}
H.~Akeb, M.~Hifi, A hybrid beam search looking-ahead algorithm
  for the circular packing problem, Journal of Combinatorial Optimization 20
  (2010) 101--130.

\bibitem{CAS}
I.~Castillo, F.~J. Kampas, J.~D. Pintér, Solving circle packing problems by
  global optimization: Numerical results and industrial applications, European
  Journal of Operational Research 191 (2008) 786 -- 802.

\bibitem{quasi2016}
D.~Zhu, Quasi-human seniority-order algorithm for unequal circles packing,
  Chaos, Solitons \& Fractals 89 (2016) 506 -- 517.

\bibitem{quasi2017}
J.~Liu, J.~Li, Z.~Lü, Y.~Xue, A quasi-human strategy-based improved basin
  filling algorithm for the orthogonal rectangular packing problem with mass
  balance constraint, Computers \& Industrial Engineering 107 (2017) 196 --
  210.

\bibitem{HE201826}
K.~He, H.~Ye, Z.~Wang, J.~Liu, An efficient quasi-physical quasi-human
  algorithm for packing equal circles in a circular container, Computers \&
  Operations Research 92 (2018) 26 -- 36.

\bibitem{HE201567}
K.~He, M.~Huang, C.~Yang, An action-space-based global optimization algorithm
  for packing circles into a square container, Computers \& Operations Research
  58 (2015) 67 -- 74.

\bibitem{He2016PackingUC}
K.~He, M.~Dosh, S.~Zou, Packing unequal circles into a square container by
  partitioning narrow action, spaces and circle items, CoRR abs/1701.00541
  (2017).

\bibitem{ZHANG20051941}
D.~Zhang, A.~Deng, An effective hybrid algorithm for the problem of packing
  circles into a larger containing circle, Computers \& Operations Research 32
  (2005) 1941 -- 1951.

\bibitem{glover1989tabu}
F.~Glover, Tabu search—part 1, ORSA Journal on computing 1 (1989) 190--206.

\bibitem{glover1990tabu}
F.~Glover, Tabu search—part ii, ORSA Journal on computing 2 (1990) 4--32.

\bibitem{ZENG2018196}
Z.~Zeng, X.~Yu, K.~He, Z.~Fu, Adaptive tabu search and variable neighborhood
  descent for packing unequal circles into a square, Applied Soft Computing 65
  (2018) 196 -- 213.

\bibitem{LodiMV99}
A.~Lodi, S.~Martello, D.~Vigo, Heuristic and metaheuristic approaches for a
  class of two-dimensional bin packing problems, Journal on Computing 11 (1999)
  345--357.

\bibitem{BansalLS-FOCS05}
N.~Bansal, A.~Lodi, M.~Sviridenko, A tale of two dimensional bin packing, in:
  46th Annual {IEEE} Symposium on Foundations of Computer Science {(FOCS},
  2005, pp. 657--666.

\bibitem{Kirkpatrick}
S.~Kirkpatrick, C.~Gelatt, M.~Vecchi, Optimization by simulated annealing,
  Science 220 (1983) 671--80.

\bibitem{gendreau2010handbook}
M.~Gendreau, J.-Y. Potvin, et~al., Handbook of metaheuristics, Vol. 272,
  Springer International Publishing, 2010.

\bibitem{LARSSON2018103}
C.~Larsson, Chapter 5 - optimization techniques, Academic Press, 2018, pp. 103
  -- 122.

\bibitem{packomania}
E.~Specht, Packomania website 2018 {www.packomania.com} (2018).

\end{thebibliography}

\end{document}